\documentclass[runningheads]{llncs}
\usepackage[T1]{fontenc}
\usepackage{graphicx}
\usepackage{times}
\usepackage{a4,url}
\usepackage{amsmath}
\usepackage{epsfig}

\newcommand{\deh} {\boldsymbol{\hat{\delta}}}%
\title{Between accurate prediction and poor decision making: the AI/ML gap (position paper)}

\author{ Gianluca Bontempi}
\institute{Machine Learning Group \\ Universit\'{e} Libre de Bruxelles\\
Bruxelles, Belgium\\
\url{mlg.ulb.ac.be}
}
 
\date{}

\newcommand{\VAR}[1]{\text{Var}\left[#1 \right]}

\newcommand{\NIL}[1]{}

\newcommand{\x}{\mathbf x}

\newcommand{\y}{\mathbf y}

\newcommand{\z}{\mathbf z}

\begin{document}

\maketitle 
\thispagestyle{empty}
\begin{abstract}
%
%
Intelligent agents rely on AI/ML functionalities to predict the consequence of possible actions and optimise the policy.
However, the effort of the research community in addressing prediction accuracy has been so intense (and successful)
that it created the illusion that the more accurate the learner prediction (or classification) the better would have been the final decision.
Now, such an assumption is valid only if the (human or artificial)  decision maker has complete knowledge of the utility of the possible actions.
 This paper argues that AI/ML community has taken so far a too unbalanced approach  by devoting excessive attention to the estimation of the state (or target) probability  
to the detriment of accurate and reliable estimations of the utility. 
In particular, few evidence exists about the impact of a wrong utility assessment on the resulting expected utility of the decision strategy.
 This situation is creating a substantial gap between the expectations and 
 the effective impact of AI solutions, as witnessed by recent criticisms and emphasised by the regulatory legislative efforts.
This paper aims to study this gap by quantifying the sensitivity of the expected utility to the utility uncertainty and comparing it to the one due to probability estimation. Theoretical and simulated results show that  an inaccurate utility assessment may as (and sometimes) more harmful than a poor probability estimation. The final recommendation to the community is then to undertake a focus shift from a pure accuracy-driven (or obsessed) approach to a more utility-aware methodology.

\end{abstract}

\section{Introduction}
After several decades of expectations and promises, AI/ML is now  pervasive in modern IT society.
As a consequence of its growing adoption, we witness the beginning of a critical assessment (from media, society and politics)
of its effective impact on the real world and society.
It is worth noting that while most success stories refer to accurate prediction performances (e.g. image/speech/text recognition)~\cite{Jordan15Mitchell}, most failures relate
to  a sort of social disillusionment about the benefits of automatic decision making. Think for instance of the
negative social reaction to AI-bots inappropriate language, racist image labelling, inadequate medical prioritising, unfair justice decisions and discriminatory hiring~\cite{Weapons16,Slota20,Banerjee20}. 
We assist then to an increasing gap between the predictive capabilities of AI/ML algorithms and 
their capacity to produce the value so long-awaited by users and society.

Nowadays,  most  prototypical/commercial AI solutions (often in the form of apps) take some sort of decisions on the basis of some predictive functionality (typically learned from historical data).
Think of a detection module returning the  probability of a fraud on the basis of transaction records, a
portable device returning the probability of a heart attack on the basis of some physiological signal or a smart car deciding to automatically brake if a pedestrian is dangerously too close.
All such applications, though related to disparate and different domains, rely on the assumption that it is possible to predict with some sufficient accuracy some state of the world (e.g. a fraud, a heart attack or a pedestrian presence) and  take a consequent appropriate action
(e.g. block the credit card, administrate some drug or stop the car). 
All of them are instances of \emph{decision making under uncertainty}~\cite{Parmigiani2009}, i.e. making choices whose consequences are not completely predictable because the state that will affect the consequences of the actions is partially unknown.
The primary concern of an AI/ML engineer is nowadays to assess and improve prediction accuracy through lots of data, statistical insight and computational resources. Such attitude relies on the implicit assumption that the more precise will be the learner, the higher will be the reward of the selected action. 
This paper claims that such approach may dangerously underestimate another important aspect of the decision process: the utility assessment.
In order to better discuss such an issue, we refer to the framework of decision theory whose goal is to help choose among actions whose consequences cannot be completely anticipated, typically because they depend on some future or unknown state of the world.  Expected utility theory~\cite{Wald49} handles this choice by assigning a quantitative utility to each consequence, a probability to each state of the world, and then selecting an action that maximises the expected value of the resulting utility. 

It is important to realise that when AI/ML is used in a (presumed) smart application, its role is not (simply) to return an accurate prediction but, more importantly, to select the most rational decision. In this perspective, reducing the uncertainty of the probability estimation is only a part of the problem.
We argue that in the AI/ML community the problem of inaccurate utility assessment has been neglected for too long or at least addressed in a too qualitative manner. We assisted in recent years to a huge effort to reduce the error of prediction, often with enormous computational (and energy) costs and sometimes with very limited effects. This has in our opinion completely biased most of the recent literature on ML 
where over-sized learning architectures (typically deep learners with millions of parameters) are employed to gain negligible percentages of accuracy. Note that this issue does not only concern supervised learning but also reinforcement learning where, apart from few exceptions (e.g. task-agnostic RL~\cite{agnosticRL}), the reward  is supposed to be known. This paper  argues that such effort could be  pointless if not accompanied by an accurate utility assessment or if used in decision tasks where only a very rough estimation of utility is available.
In this sense, this paper advocates a \emph{frugal} approach~\cite{frugal21} to machine learning by insisting on the fact that an oversized effort to improve accuracy could be unnecessary and made vain by other aspects of the decision problems.

\section{Decision theory framework}
Decision-making under uncertainty deals with making choices whose consequences are not completely predictable, 
because the state (that will determine the consequences of the actions) is partially unknown (or latent)\footnote{Random variables are denoted by the bold notation.}. 
The seminal work of Wald~\cite{Wald49} axiomatises the decision problem by assigning a quantitative utility
to each action, a probability distribution to the state and selecting the action that maximises the expected utility.
In other terms the rational behaviour under uncertainty is postulated as a behaviour which maximises expected utility: this approach
requires the knowledge (or at least an estimation) of both the utility terms and the state probability distribution.

For the purpose of this paper, we will consider  a simple decision theory task
where the state is binary $\theta \in \{0,1\}$ and we have to take an appropriate action out of two alternatives $\{a_0,a_1\}$. 

Let us consider a simple decision-making problem

\begin{center}
\begin{tabular}{c||c|c}
Action & State $\theta=0$ & State $\theta=1$\\
\hline
$a_0$ &  $c_{00}=0$    & $c_{01}$\\
$a_1$ & $c_{10}$ & $c_{11}=0$\\
\end{tabular}
\end{center}

where $c_{01}$ ($c_{10}$) denotes the cost incurred if we take the action $a_0$ ($a_1$) when the state is $\theta=1$ ($\theta=0$).
Note that such costs boil down to the famous false positive and false negative costs encountered in any binary classification task.
To simplify the cost distribution, we  suppose that for a given state configuration it exists the "right action", i.e. the action which returns a null cost (e.g. we have no loss if an email is spam and we delete it).

This setting is simple but often representative of many smart applications where the unknown state is binary (spam/ham, fraud/genuine, health/sick, safe/unsafe) and an action (remove/keep, block/process, prescribe a treatment or not, raise alert or not) on the basis of the state detection is required.

Dealing with decision-making under uncertainty requires  the estimation of the latent state and the assessment of the utilities (or costs) of the actions for the possible state configurations.
 A rational decision maker should proceed by assigning numerical utilities to action consequences and scoring actions by their expected utility
 where the expectation is obtained by estimating the probability of the state configurations. So if we know that
 $P({\boldsymbol \theta}=0)=p_0$ (and consequently $P({\boldsymbol \theta}=1)=1-p_0$) the optimal (also called Bayes)  action  is the one that minimises the expected loss score
 (or maximises the expected utility)
 $$
 a^*=\arg \min_{a_0,a_1} E[C(a,{\boldsymbol \theta})]=
 \begin{cases}
 a_0 & \mbox{ if } c_{01} (1-p_0) < c_{10} p_0\\
 a_1 & \mbox{ else}
 \end{cases}
 $$
 Note that if the optimal action is chosen, the associated expected loss is
 \begin{equation}
 \label{eq:minc}
 L^*=\min(c_{01} (1-p_0), c_{10} p_0)
 \end{equation}
In an uncertain setting, unfortunately, the probability distribution of the state is not known but has to be estimated (e.g. from observed data).
Since the state is binary (e.g. spam/ham, sick/healthy) the estimation of the state configuration boils down to a problem of binary classification where observed data can be used to learn a model returning the conditional probability of the state given some features.
In a practical setting, this is not the only unknown quantity, since  $c_{01}$ and $c_{10}$  need to be estimated as well.
The difficulty in assessing utility terms in a quantitative manner is well-known in the decision theory community.
The elicitation of reliable utilities is one of the major obstacles in the application of  decision  analysis  to  real  problems.  
This is particularly studied in the medical context where the definition of utility (e.g. what is the cost of not detecting a heart attack)
is particularly hard and depends on a lot of disparate factors (ethical, religious, subjective, time-dependent, relative vs absolute risk)~\cite{Chapman00}.
This paper does not propose a solution to this problem but aims to quantify (and compare) the sensitivity of the  expected loss to the uncertainty
of probability and utilities, respectively.

\subsection{Uncertainty and suboptimal action}
\label{sec:unce}
In a real setting, we do not have access to the parameters describing the decision task but we can only estimate them (e.g. 
by means of a machine learning procedure). As a consequence, we won't necessarily be able to select the optimal action.
In what follows we will analyse the impact of (probability and utility) estimation errors on the suboptimality of the action selection and the consequent error increase.

Let us introduce first the  quantity
\begin{equation}
\label{eq:delta}
{\delta}=  c_{01} (1-p_0) -  c_{10} p_0, 
\end{equation}
whose absolute value denotes the loss increase in the case of a suboptimal choice.
Consider also its estimate  
\begin{equation}
\label{eq:deltahat}
 \hat{\delta}=  \hat{c}_{01} (1-\hat{p}_0) -  \hat{c}_{10} \hat{p}_0
\end{equation}
which depends on $\hat{p}_0$, $\hat{c}_{01} $ and $\hat{c}_{10}$, that are the estimates   of the state probability distribution and action utilities, respectively.
It follows that the selected action in an uncertain setting will depend on $\hat{\delta}$  and will be
$$
 \hat{a}=
 \begin{cases}
 a_0 & \mbox{ if } \hat{\delta} < 0\\
 a_1 & \mbox{ else}
 \end{cases}
 $$
Note that the action selection is not optimal ($a^* \neq \hat{a}$)  when the signs of $\delta$ and $\hat{\delta}$ are discordant. 
In such cases, we will incur an  expected loss increased by the quantity $|\delta|$.

Suppose that the estimators  $\hat{{\mathbf c}}_{01}, \hat{{\mathbf c}}_{10}, \hat{{\mathbf p}}_0$ are unbiased yet affected by the variances
$\sigma^2_{01}$, $\sigma^2_{10}$ and $\sigma^2_{p_0}$, respectively. Also, for the sake of simplicity, we will assume that the distribution
of $\deh$  is normal and with mean $\delta$.
It is possible then to obtain an analytical expression of the increase of the expected loss. First of all, let us compute the probability $P_{err}$ of
making an error in the action selection. The probability of an erroneous selection is
$P(\deh <0) $ when $\delta>0$ and $P(\deh >0) $ when $\delta<0$. If we assume that $\deh$ is an unbiased estimator of 
$\delta$ and $\delta>0$, the probability of an erroneous selection boils down to the probability that a normal random variable with a positive mean takes negative values\footnote{Note that by symmetry the result if identical if $\delta<0$.}.
For  a normal r.v. $\z$ with mean $\mu$ and variance $\sigma^2$
$$P( \z \le z)=\Phi\left(\frac{z-\mu}{\sigma} \right)= \frac{1}{2}\left[1+\mbox{erf} \left( \frac{z-\mu}{ \sqrt{2\sigma^2}}\right) \right]
$$
where $\Phi$ is the normal cumulative function and erf$(\cdot)$ is the Gauss error function.
Then it follows
\begin{equation}
\label{eq:perr}
P_{err}=P(\deh <0 | \delta>0) = \frac{1}{2}\left[1+\mbox{erf} \left( \frac{-|\delta|}{ \sqrt{2\hat{\sigma}^2}}\right) \right] 
\end{equation}
where $\hat{\sigma}^2$  is the variance of $\deh$.

Since the variance of the product of two independent variables $\x$ and $\y$ is
$$\VAR{\x \y}=\VAR{\x } \VAR{\y} +\VAR{\x} (E[\y])^2+ \VAR{\y} (E[\x])^2 $$
we obtain

\begin{multline}
\label{eq:vard}
\VAR{\deh}=\hat{\sigma}^2=\VAR{\hat{{\mathbf c}}_{01}} \VAR{\hat{{\mathbf p}}_0} +\VAR{\hat{{\mathbf c}}_{01}} (1-p_0)^2+ 
\VAR{\hat{{\mathbf p}}_0} (c_{01})^2 +\\
\VAR{\hat{{\mathbf c}}_{10}} \VAR{\hat{{\mathbf p}}_0} +\VAR{\hat{{\mathbf c}}_{10} }(p_0)^2+ \VAR{\hat{{\mathbf p}}_0} (c_{10})^2 
\end{multline}

Now since each time an error is committed the increased loss amounts to $|\delta|$,
it follows from~\eqref{eq:perr} that the expected increase amounts to

\begin{equation}
\label{eq:Delta}
\Delta= P_{err} |\delta| = \frac{|\delta|}{2}\left[1+\mbox{erf} \left( \frac{-|\delta|}{ \sqrt{2\hat{\sigma}^2}}\right) \right]
\end{equation}
where $\delta$ and $\hat{\sigma}^2$ are defined in~\eqref{eq:vard} and~\eqref{eq:delta} respectively.
This equation formalises the relationship between the increase in the expected loss and the  estimation errors
of both the probability of the latent state and the utility terms. By plotting such relation it is possible  to visualise i) the sensitivity of
the expected loss increase to the probability estimation error  ii) the sensitivity of
the expected loss increase to the utility estimation error and iii) the joint effect due to errors in both probability and utility terms. Note also that the 
expression~\eqref{eq:vard} could be easily extended to the case where the costs $c_{00}$ and $c_{11}$ are not null. 

\subsection{Sensitivity study}

This section visualises the relation between estimation errors and expected loss increase $\Delta$ due to a suboptimal action selection
on the basis of the analytical formula~\eqref{eq:Delta}. In particular, we focus on distinguishing the impact of errors in probability estimation vs  utility estimation.

\begin{figure}
 \centering
 \includegraphics[angle=0,scale=.45]{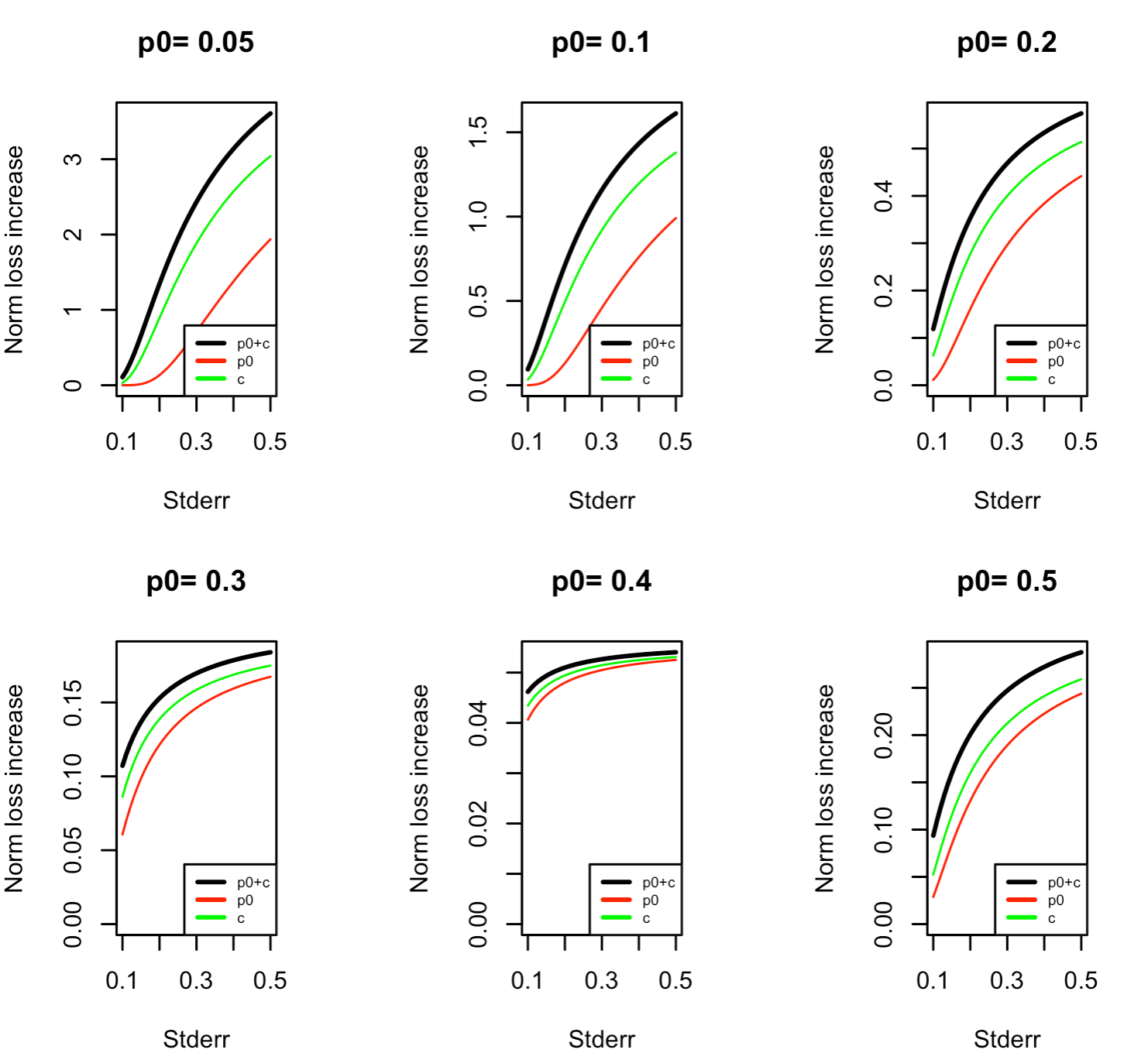}
 \caption{Normalised increase $\Delta/L^*$ of the expected loss for different values of $p_0$, $c_{01}=0.3$ and $c_{10}=0.5$ 
 and different standard errors. 
 The green curve represents the normalised expected increase due to the estimation error of the cost terms.
 The red curve represents the normalised expected increase due to the estimation error of the probability distribution.
The black curve represents the normalised expected increase due to the estimation errors of both  probability and costs.
  \label{fig:ecost}}
\end{figure}

We consider the set of values $p_0 \in \{0.05, 0.1, 0.2, 0.3, 0.4, 0.5 \}$, $c_{01}=0.3$ and $c_{10}=0.5$ and a set of  increasing values of  standard errors (i.e. standard deviation of the estimator). 
Figure~\ref{fig:ecost} plots the  normalised increase $\Delta/L^*$ where $L^*$ denotes the minimal loss~\eqref{eq:minc} for different values of standard errors\footnote{R script at \url{https://tinyurl.com/4yes8jda}}. Note that in the green curves the standard error $\sqrt{\VAR{\hat{{\mathbf p}}_0}}=0$ since
we consider only uncertain utilities while in the red case $\sqrt{\VAR{\hat{{\mathbf c}}_{10} }}=\sqrt{\VAR{\hat{{\mathbf c}}_{01} }}=0$ since the utility terms are supposed to be perfectly known. In the black
curve $\sqrt{\VAR{\hat{{\mathbf p}}_0}}=\sqrt{\VAR{\hat{{\mathbf c}}_{10} }}=\sqrt{\VAR{\hat{{\mathbf c}}_{01} }} >0$ since we assess the impact of both uncertainties.

As expected the loss increase is positively correlated with the standard error.
Also, it appears that the normalised $\Delta$ can be more sensitive to the cost (utility) estimation error (green curve) than to the probability estimation (red curve). This is essentially due to the fact that being a probability a bounded quantity, so is its variance. As a consequence, a limited focus to the probability estimation may be vain if cost terms are inaccurate. The figure shows that in some specific configurations (typically low entropy of the probability distribution) an error in the estimation of the cost terms  may lead to a final expected loss which is three times higher than the optimal one. This means that even a very accurate estimator of the state probability may lead to a strategy which is very far from the optimum if insufficient attention is devoted to the assessment of the utility.

\section{Experiments}
This section aims to assess the quality of the analytical approximation~\eqref{eq:Delta} by performing a number of simulated
experiments with 25 different values of $p_0$ uniformly spread over $[0,1]$, 25 uniformly random values of $c_{01} \in [0.2,0.4]$ and $c_{10} \in [0.4,0.6]$ and relaxing the normal and independency assumptions made in Section~\ref{sec:unce}. 
We assume that the probability and the cost estimators have a Beta and Uniform distribution, respectively, and that the costs are constrained to be positive. 

\begin{figure}
 \centering
 \includegraphics[angle=0,scale=.4]{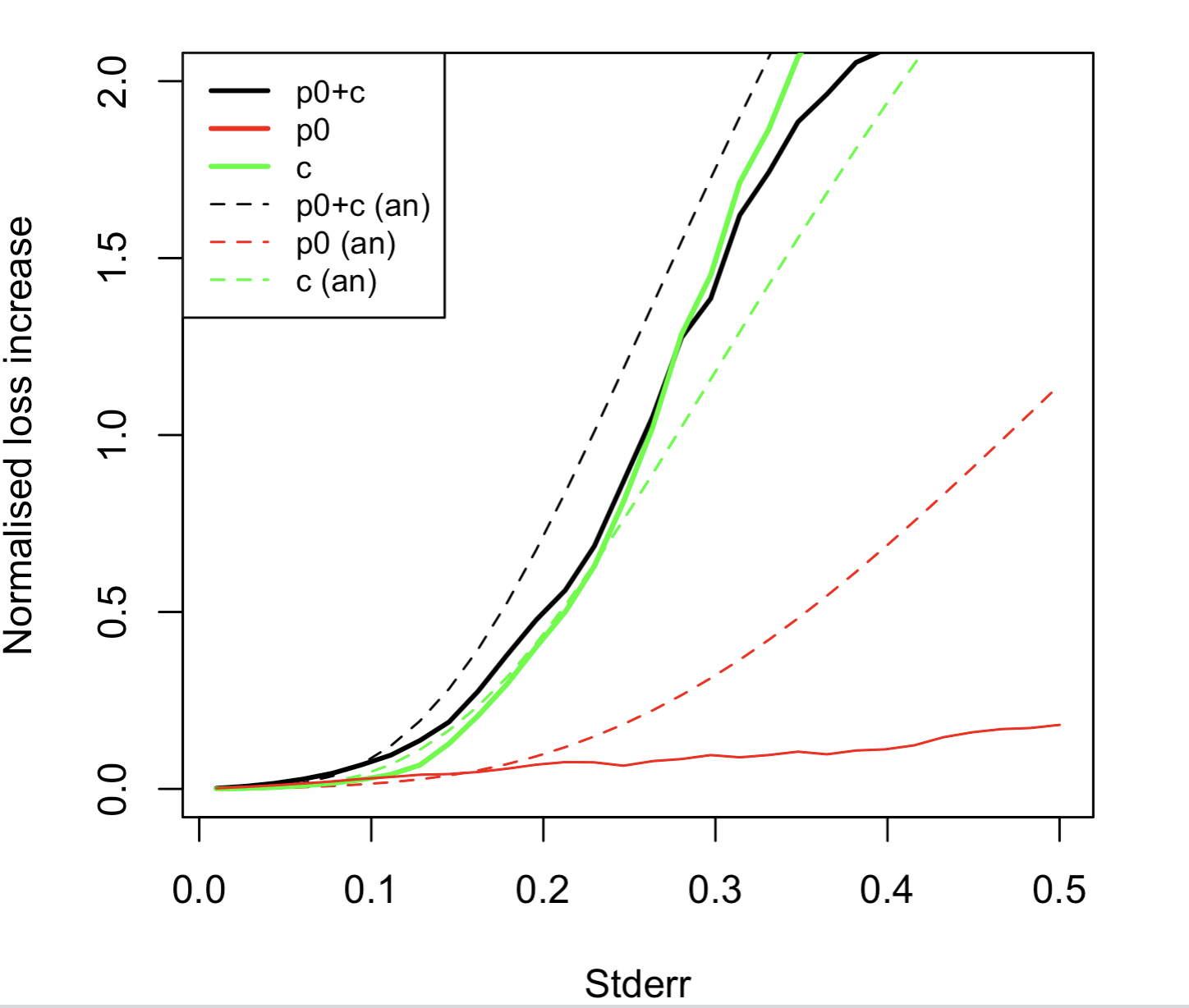}
 \caption{Average normalised increase $\Delta/L^*$ of the expected loss for different standard errors. 
 The green curve represents the normalised expected increase due to the estimation error of the cost terms.
 The red curve represents the normalised expected increase due to the estimation error of the probability distribution.
The black curve represents the normalised expected increase due to the estimation errors of both  probability and costs.
Solid curves refer to simulated values under non-normal assumptions. Dashed lines are obtained by the analytical approximation~\eqref{eq:Delta}.
\label{fig:scost}}
\end{figure}

Figure~\ref{fig:scost} shows the normalised loss increase (averaged over all the possible combinations of  $p_0$, $c_{01}$ and $c_{10}$ values)
vs. the standard errors of the estimators. The meaning of the red, green and black curves in terms of associated standard error is
the same as in Section~\ref{sec:unce}. 

It is interesting to see in Figure~\ref{fig:scost} that the analytical approximation is somewhat faithful over a large number of
experimental configurations. Also, the higher sensitivity of the expected loss to the cost uncertainty is largely confirmed by the simulated experiment\footnote{R script at \url{https://tinyurl.com/ye2y369s}}.

\section{Discussion and conclusion}
Nowadays there is an exploding adoption in smart devices of AI/ML architectures involving millions of parameters (and generating tons of emissions) 
because of their claimed gain in accuracy. Nevertheless, we rarely ask ourselves what we really know about the cost of a false positive or a false negative.
Since the goal of an ML predictor is not (only) the prediction (e.g. no doctor is interested in the exact probability of a heart attack) but the 
consequent decision making (e.g. should the patient take a drug now), we showed that most of the effort done in designing more and more complex learners could be made vain by the ignorance about the decision utilities.
The complexity of assessing reliable utilities is well known in the medical world. However, nowadays the adoption of automated decision-making goes largely beyond such a domain and requires the adoption of sound techniques to  assess the uncertainty of the
utility terms. This does not necessarily mean that estimating utility is an easy or feasible task\footnote{See our discussion on incommensurability and ethics in \mbox{\url{https://datascience741.wordpress.com/2018/08/}}}. Nevertheless, neglecting its importance boils down to 
miss an important side of the problem.
As stressed by Bernoulli in 1738, "no valid measurement of the value of risk can be given without consideration of its utility".
An interesting direction is represented by the recent work on agnostic reinforcement learning\footnote{\url{https://tarl2019.github.io}} though exploration phases are not always allowed in realistic tasks and the distribution of the utility function is not necessarily available. The position of this paper is that a boost of AI/ML research on the assessment of reliable utility from data is definitely necessary. The same effort devoted to the predictive task should be addressed to characterise the nature of the utility function in terms of uncertainty, relevant features (e.g. different users or working conditions could require different utilities) and stationarity (e.g. evolution of the cost function with time).  Also, we personally consider that this would be a much more worthy task for an AI scientist than spending her time optimising a million parameters  neural network...
\subsection*{Acknowledgments}
Gianluca Bontempi is supported by the Service Public de Wallonie Recherche under grant nr 2010235-ARIAC by DigitalWallonia4.ai. Computational resources have been provided by the Consortium des Equipements de Calcul Intensif (CECI), funded by the Fonds de la Recherche Scientifique de Belgique (F.R.S.-FNRS) under Grant No. 2.5020.11 and by the Walloon Region. Gianluca Bontempi thanks Théo Verhelst and Gianmarco Paldino for useful comments and remarks.
\bibliography{benelearn}
\end{document}